\documentclass{amsart}
\usepackage{amsthm}
\usepackage{amsmath}
\usepackage{amsfonts}
\usepackage{amssymb}
\usepackage[hidelinks]{hyperref}
\usepackage[ruled,linesnumbered]{algorithm2e}
\usepackage[noend]{algpseudocode}
\usepackage{listings}
\usepackage{graphicx}
\usepackage{subfig}
\usepackage{chngcntr}
\counterwithin{figure}{section}
\counterwithin{table}{section}
\usepackage{float}

\floatstyle{ruled}
\newfloat{code-listing}{thp}{lop}
\floatname{code-listing}{Code Listing}

\def\real{{\mathbb{R}}}

\def\expectation{{\mathbb{E}}}
\def\Xs{\mathcal{X}}
\def\Ys{\mathcal{Y}}
\DeclareMathOperator{\sign}{sign}
\DeclareMathOperator{\sig}{sig}
\begin{document}
\title[DistFom \& Spark]{\large Distributed Function Minimization in Apache Spark}
\author{Andrea Schioppa}
\address{Amsterdam, Noord Holland}
\email{ahisamuddatiirena+math@gmail.com}
\begin{abstract}
We report on an open-source implementation
for distributed function minimization on top of
Apache Spark by using gradient
and quasi-Newton methods.
We show-case it with an application to Optimal Transport
and some scalability tests on classification and regression problems.
\end{abstract}
\maketitle
\tableofcontents
\lstset{
language=Scala,
keywordstyle=\bfseries\itshape,
identifierstyle=\scshape,
mathescape,
}
    \section{Introduction}\label{sec:intro}
\subsection{Motivation}
\label{sec:motivation}
In this paper we report on an open-source implementation
for distributed function minimization using gradient
and quasi-Newton methods.
\par This library \lstinline{distfom} (GitHub \href{https://github.com/salayatana66/distfom}{repo}
\footnote{https://github.com/salayatana66/distfom}) is
geared towards mathematical problems that involve minimization
of functions whose value / gradient require costly computations.
The original motivation was to distribute computations related
to Optimal Transport~\cite{trigila_thesis} where the underlying feature
spaces have high cardinality and large amount of data-points
are required in the data-driven approach.
\par A possible approach to distribute such calculations might
be renting a big server or using dedicated clusters with
fast network connections relying on a message-passing standard
for parallel computing, e.g.~MPI.
However, we worked on a solution for ``cheap'' commodity clusters that,
for example, can be rented from a Cloud provider or shared
among users in a proprietary e-commerce company cluster.
\par In this shared setting, besides the usual problem of node
failure, we wanted to address a solution that can work with
preemptible nodes, i.e.~nodes that can be claimed back
from the cloud provider or a job scheduler (e.g.~in the
e-commerce company scenario depending on the priority queue of the job).
Being able to use preemptible nodes is often
more cost effective than renting a locked resource.
For example, at the time of writing a preemptible node is 80\% cheaper on
the Google Cloud Platform.
\subsection{Extension to Machine Learning}
\label{sec:ml-ext}
\par From a Machine Learning perspective a large
distributed Optimal Transport Problem presents both the
challenges of \emph{model parallelism}, as one needs
the value of the potentials at each sampled point, and of
\emph{data parallelism}, as the cost matrix is large
and has to be stored in a distributed file system.
\par Therefore, using the same library for ML problems is
viable.
We thus implemented linear models and Factorization Machines, and
loss functions for regression, (multi)-classification and ranking.
\par Another motivation for this extension was this
\href{https://booking.ai/crunching-big-data-with-4-machine-learning-libraries-284ae3167885}{blog post}
\footnote{\texttt{https://booking.ai/crunching-big-data-with-4-machine-learning-libraries-284ae3167885}}
which we co-authored.
The topic of the post is a comparison of open-source ML libraries
for a regression problem relevant to an e-commerce company.
We found using H2O slightly unsatisfactory because of limitations
in handling categorical features with high-cardinality and
the cost of requiring non-preemtible nodes.
We found Vowpal Wabbit~\cite{langford-hashing, langford-all-reduce, vowpal-wabbit}
competitive both in a single
and multi-node setting with the all reduce implementation~\cite{langford-all-reduce}.
However, Vowpal Wabbit requires broadcasting the whole model weights
to each worker, thus limiting the cardinality of feature interactions
by having to choose a smaller value of the hashing parameter.
Finally, Spark-ML gave very good results out of the box and
was very efficient relying on \lstinline{breeze}~\cite{breeze}
using LBFGS for function-minimization instead of Stochastic Gradient
Descent.
However, also for Spark-ML we had to bucketize feature interactions
as model weights had to be broadcasted to each worker.
While looking for solutions to shard the model weights across
executors we came across the master thesis~\cite{ehsan-thesis} which
used Spark to solve large convex linear problems that require distributing
vectors and matrices across executors.
\subsection{Choice of Spark}
\label{sec:spark-choice}
We decided to build the library on top of an
already existing / well-established framework.
Note also that we focused on building a \emph{library}
rather than \emph{application} for ease of future
extendability.
This is in contrast to the design
choices of Vowpal Wabbit which, having virtually just
one external dependency, manages to be intrinsically
very fast at optimized but, as we feel, at the cost of
extendability and integration with other systems.
\par A popular platform is Hadoop~\cite{hadoop}, an open-source implementation
of MapReduce~\cite{mapreduce}; as observed by
several authors, e.g.~\cite{zaharia-rdd,langford-all-reduce,ehsan-thesis}
the MapReduce
framework does not fit well with iterative algorithms
as it exposes a somewhat limited programming model and
incurs repeated I/O operations not making full-usage
of data caching.
\par These issues were addressed by Spark~\cite{zaharia-rdd,spark} which exposes
a rather expressive programming model based on RDDs in
a framework that can be run on top of several resource managers
like Kubernetes, Mesos and Yarn.
\par Spark takes care for the user of fault-tolerance,
speculation to overcome presence of slow machines and
usage of preemptible workers.
Moreover one can develop in Scala~\cite{scala} having access to a rich
ecosystem of libraries written in Java and / or Scala.
\par We thus decided to use Spark, being however aware
of a disadvantage: it is not clear to us how to run
an implementation of a \emph{Parameter Server}~\cite{li-param-server, smola-first-par-server}
on top of the Spark library.
The Parameter Server~\cite{li-param-server, smola-first-par-server} is a very powerful framework which
allows to combine
data-parallelism and model-parallelism with very
favorable results, compare~\cite{li-param-server}.
Therefore we end up using a different computational model than
Parameter Server.
However we just mention in passing of Tencent's
Angel~\cite{Angel}, an open-source implementation of Parameter Server which
has a module Spark \emph{on} Angel.
\subsection{Previous Work}
\label{subsec:prev-work}
\par The Scala library \lstinline{breeze}~\cite{breeze} implements
several numerical algorithms;
in particular it contains a general and extensible framework
for function minimization based on gradient descent and quasi-Netwon
methods.
The architecture of our \lstinline{FirstOrderMinimizer} is
based on the corresponding one in the \lstinline{breeze} package.
\par In~\cite{ehsan-thesis} the author implemented Mehrotra's predictor-corrector
interior point algorithm on top of Spark to solve large linear programming
problems;
his implementation requires distributing vectors and matrices.
\par For a nice overview of work about ML frameworks implementing
parallel computing we refer to~\cite{li-param-server}.
Here we just mention three approaches:
\begin{itemize}
    \item Vowpal Wabbit with AllReduce: in~\cite{langford-all-reduce} the authors extend Vowpal Wabbit
    to a multi-node setting using AllReduce on top of Hadoop.
    They report on fitting a click-through model via a logistic regression;
    the model having $2^{24}\sim 16.8M$ parameters and their cluster using
    about 1000 nodes.
    This framework is very fast and efficient but it only implements \emph{data parallelism}
    as model parameters need to be broad-casted to all the nodes.
    A very nice point made in~\cite{langford-all-reduce} is the mixing of gradient descent for
    parameter initialization with LBFGS for fast convergence.
    \item GraphLab~\cite{distro-graphlab,graph-lab}: it is a graph-parallel framework to express computational
    dependencies.
    In~\cite{distro-graphlab} a thorough comparison is made between GraphLab, Hadoop and
    MPI.
    The comparison looks very favorable to GraphLab in terms of resource usage.
    While part of it might be ascribed to usage of C++ instead of Java (Hadoop
    is written in Java), improvements are likely driven by asynchronous
    communication between nodes.
    \item Parameter Server: as observed in~\cite{li-param-server} GraphLab lacks elastic
    scalability and is impeded by relying on coarse grained snapshots for
    fault tolerance.
    Parameter Server's paradigm can be used to combine the elasticity of
    Spark or Hadoop with asynchronous updates.
    Instead of just relying on worker machines to process the training data,
    a few machines, called the servers, hold model weights, send them to
    workers processing training examples and update them asynchronously.
    An open-source implementation of Parameter Server which can be run on top
    of Yarn is Tencent's Angel.
    \item From the Optimal Transport literature we cite the parallel
    computation of Wasserstein barycenters~\cite{uribe-wasserstein, parallel-wasserstein}
    and the large scale computation
    of Wasserstein distances~\cite{genevay_large_opt}.
    For our setting~\cite{genevay_large_opt} is particularly relevant as gradient descent is used
    for computing Sinkhorn distances between distributions of 20k sampled
    points from a space of word-embeddings.
    In~\cite{genevay_large_opt} computations are sped up by using 4 Tesla K80 GPUs.
\end{itemize}
\subsection{Contributions}
\label{subsec:contributions}
\par We implemented an open-source library for distributed
function minimization built on top of Spark and consistent with the
\lstinline{breeze} package.
\par We implemented minimization of the optimal transport loss
with entropic regularization~\cite[Chap.~4]{cuturi_book}.
We also implemented linear models, Factorization Machines~\cite{rendle-bpr, rendle-fm} and commonly
used losses for supervised learning and ranking problems.
\par From a computational approach we implement a mixing of data and
parameters splitting that allows circumventing the implementation of a
Parameter Server in Spark.
\subsection{Future work}
\label{subsec:future-work}
\par Arguably a favorable way to minimize a function in a setting
where one can compute gradients globally or on large batches of
data benefits from quasi-Newton or second order methods.
\par However, as observed in~\cite{langford-all-reduce} proper initialization is critical;
convergence speed and algorithmic stability usually require being in a
neighborhood of a solution.
While for the optimal transport problem initialization is straightforward
(with non-negative costs just set all the potentials to $0$), for
ML problems it can be tricky.
The work~\cite{langford-all-reduce} recommends using gradient descent on mini-batches to
initialize model parameters before using a quasi-Newton method.
In a model-parallel setting this can be relatively tricky and we
leave its implementation for future work.
However, in the current implementation one might already proceed with
Adaptative Gradient Descent on larger batches.
\par We finally leave for future work an implementation of Parameter Server.
    \section{Architecture and Implementation}\label{sec:arch-impl}
    \subsection{High Level view of the Architecture}\label{subsec:high-level-view}
Let us look at how one might want to implement a First Order Minimization
algorithm in a distributed setting.
Such an algorithm, which we represent as an \lstinline{abstract class FirstOrderMinimizer}
(abr.~\lstinline{FoMin}) takes as input a differentiable function
$f$ and an initial point $w_0$.
Then the algorithm starts updating $w_0$ generating a sequence $\{w_k\}_k$;
during this process \lstinline{FoMin} keeps track of
$f(w_k)$, $\|\nabla f(w_k)\|$ to decide when to stop.
\par Now $f$ is an object of which we must be able to compute values and
gradients.
In some cases, for the sake of computational efficiency,
one might want different implementations depending on whether the value
or the gradient of $f$ are required, and hence we define differentiable functions as a
\lstinline{trait} having methods \lstinline{computeValue($w$: $T$): $F$},
\lstinline[mathescape]{computeGrad($w$: $T$): $T$} and \lstinline[mathescape]{compute($w$: $T$): $(T, F)$}
where \lstinline[mathescape]{$T$, $F$} are the types of $w$ and $f(w)$ (i.e.~the scalar field),
respectively.
\par When $f$ is the \emph{empirical loss} associated to a Machine Learning model,
the value of $f$ will also depend on data $\mathcal{D}$, while $w$ will usually
represent model parameters.
In this case it might be burdensome to evaluate $f$ using all the data
$\mathcal{D}$ and one might opt for a partitioning
$\mathcal{D}=[\mathcal{B}_0,\cdots,\mathcal{B}_{m-1}]$ into $m$ batches.
To be clear, here we expect the batches to be \emph{big} contrary to the case
of the batches considered in training models using Stochastic Gradient
Descent.
Then at each evaluation of $f$ we might want to rotate the data batch
used for evaluation, so that at the $n$-th evaluation the
batch $\mathcal{B}_{n\mod m}$ is used.
In some cases (e.g.~during line searches) we might however want to stop such a batch rotation and
that is why the \lstinline{trait DistributedDiffFunction} (abr.~\lstinline{DDF}) has
methods \lstinline{holdBatch: $()$} and \lstinline{stopHoldingBatch: $()$}.
\par The argument of a \lstinline{DDF} must be a vector amenable to operations
needed by gradient descent algorithms.
We thus introduce the \lstinline{trait FomDistVec} which besides the
operations \lstinline{+,-,*,/} has:
\begin{itemize}
    \item \lstinline{dot(other: $FomDistVec$) : $F$} taking the dot
    product of \lstinline{this} and \lstinline{other}, as for example
    required by LBFGS.
    \item \lstinline{norm(p: $Double$): $F$} representing the $l_p$
    norm of \lstinline{this}, as required when computing regularization
    or checking a convergence condition requiring $\|\nabla f\|$ to be
    small.
    \item \lstinline{keyedPairWiseFun[$K$](b: $FomDistVec$)(f: $K$): $FomDistVec$}
    which maps \lstinline{this}, \lstinline{b} to a vector having as
    component $i$ the value of \lstinline{f($i$, this($i$), b($i$))}, which is
    required by some algorithms involving $l_1$ regularization, as for example
    OWLQN.
\end{itemize}
We also make \lstinline{FomDistVec} extend:
\begin{itemize}
\item \lstinline{Persistable} to deal with caching and un-caching vectors.
    \item \lstinline{InterruptLineage} to allow to truncate long \lstinline{RDD}
    lineages that are generated by iterative algorithms.
    Indeed, with multiple iterations lineages can get long and stall the
    scheduling of tasks.
\end{itemize}
Finally, the backbone of each implementation of \lstinline{FomDistVec} is
an implementation of the \lstinline{trait DistVec} which supports
just the basic component-wise operations \lstinline{+,-,*,/}
besides extending \lstinline{Persistable}.
    \subsection{Implementations of {\normalfont\lstinline{DistVec}}}\label{subsec:dist-vecs}
We can now concretely discuss implementations of distributed vector types
starting from the simplest \lstinline{trait DistVec}.
The basic idea is to split a vector $w$ of length $e$ (with indexing starting
at $0$) into $\frac{e}{eb}$ blocks where $eb$ is the number of elements per
block.
Concretely each block is a partition of an \lstinline{RDD[$(Int, V)$]}
containing a single element $(i, v)$, and the $j$-th component of
$v$ would then represent the element of $w$ with index $i*eb + j$.
Then pairwise vector operations can be implemented efficiently
using \lstinline{zipPartitions};
to further make the implementation stable we make the partitioning
be enforced via a \lstinline{Partitioner} of type \lstinline{BlockPartitioner}.
\par Two complications should be added to this picture.
The first is that we deal with two vector types \lstinline{$V$}:
\begin{itemize}
    \item \lstinline{DenseVector[$F$]} from the \lstinline{breeze} library
    representing a standard \emph{dense} vector.
    The resulting implementation of \lstinline{DistVec} is named
    \lstinline{DistributedDenseVector[$F$]} (abb.~\lstinline{DDV[$F$]}).
    \item \lstinline{DenseMatrix[$F$]} to represent a family of vectors
    $\{w_\alpha\}_{\alpha=1}^m$ where $m$ is \emph{small} and each
    partition now contains \lstinline{($i$, $M$)} where $M$ is an
    $m\times eb$-matrix.
    We think of these vectors as being distributed across the column
    dimension and stacked on top of each other across the row dimension.
    Such vectors arise when we want to implement algorithms dealing with
    multi-label classification or Factorization Machines models.
    For this reason we name the resulting implementation
    \lstinline{DistributedStackedDenseVectors[$F$]} (abb.~\lstinline{DSDV[$F$]}).
\end{itemize}
Secondly we use \emph{generics} to represent the field \lstinline{$F$},
with specialized implementations of \lstinline{$Double$}, \lstinline{$Float$}.
We do not want to dwell on this topic, but mention in passing that we take
advantage of Scala features like \emph{implicit values}\footnote{\texttt{https://docs.scala-lang.org/tour/implicit-parameters.html}}
(cmp.~\lstinline{GenericField.scala})
in the \href{https://github.com/salayatana66/distfom}{GitHub Repo}\footnote{https://github.com/salayatana66/distfom} and reflection using
\lstinline{ClassTag}\footnote{\texttt{https://www.scala-lang.org/api/2.12.3/scala/reflect/ClassTag.html}}
and \lstinline{TypeTag}\footnote{\texttt{https://docs.scala-lang.org/overviews/reflection/typetags-manifests.html}}.
\subsection{Implementations of {\normalfont \lstinline{FomDistVec}}}\label{subsec:fomdist-impl}
Here we provide two implementations of \lstinline{FomDistVec}:
\lstinline{FomDistDenseVec} (abr.~\lstinline{FDDV}) built on top of
\lstinline{DDV} and \lstinline{FomDistStackedDenseVec} (abr.~\lstinline{FDSDV}) built on top
of \lstinline{DSDV}.
The implementation is straightforward;
we just note that to compute dot products and norms we use the method
\lstinline{treeAggregate} provided by Spark in the spirit of the
``AllReduce'' mentioned in~\cite{langford-all-reduce}.
On the language side, for the sake of interoperability with \lstinline{DDF} both
\lstinline{FDDV} and \lstinline{FDSDV} must be castable back and forth
to \lstinline{FomDistVec}.
The Scala language offers an easy solution via \emph{implicit conversions}\footnote{\texttt{https://docs.scala-lang.org/tour/implicit-conversions.html}}.
\subsection{Architecture of {\normalfont\lstinline{FirstOrderMinimizer}}}\label{subsec:fom-architecture}
The architecture of \lstinline{FoMin} is based on the \lstinline{breeze} package.
The basic idea is that the most common iterative minimization algorithms using function
values and gradients can be described in the same framework, see the following code:

\begin{lstlisting}[frame=shadowbox]
def minimizeAndReturnState($\tilde f$: $DDF$, init: $T$): State = {
// adjust the objective, e.g. if $l_1$-regularization is used
val $f$ = this.adjustFunction($\tilde f$)

// state initialization
var numSteps = 0
var $x$ = init
var $h$ = this.initialHistory($f$, $x$)
var ($f(x)$, $\nabla f(x)$) = this.calculateObjective($f$, $x$, $h$)
var state = new State($x$, value = $f(x)$, grad = $\nabla f(x)$,
                      history = $h$,
         convergenceInfo: Option[$ConvergenceReason$] = None)

// the iteration loop
while(state.convergenceInfo.isEmpty) {

    // compute descent direction
    val $w$ = this.chooseDescentDirection(state, $f$)
    $w$.persist()
    if ((numSteps > 0) & (numSteps % this.interruptLinSteps
                          == 0))
          $w$.interruptLineage()
    $w$.count()

    // compute the step size
    if(this.holdBatch) $f$.holdBatch()
    val $\eta$ = this.determineStepSize(state, $f$, $w$)
    if(this.holdBatch) $f$.stopHoldingBatch()

    // update $x$
    $x$ = this.takeStep(state, $w$, $\eta$)
    $x$.persist()
    if ((numSteps > 0) & (numSteps % this.interruptLinSteps
                          == 0))
          $x$.interruptLineage()
    $x$.count()

    // free storage for $w$
    $w$.unpersist()

    // compute new objective and gradient
    ($f(x)$, $\nabla f(x)$) = this.calculateObjective($f$, $x$,
                                   state.history)
    $\nabla f(x)$.persist()
    if ((numSteps > 0) & (numSteps % this.interruptLinSteps
                          == 0))
          $\nabla f(x)$.interruptLineage()
    $\nabla f(x)$.count()

    // measure improvement of objective
    val impr = ($|$state.value - $f(x)$$|$)/($max($state.value $, 10^{-6})$)
    println(s"Relative improvement: $\$$impr")

    // update history and check convergence
    $h$ = this.updateHistory($x$, $\nabla f(x)$, $x$, $f$, state)
    val newCInfo = this.convergenceCheck($x$, $\nabla f(x)$, $f(x)$,
                state, state.convergenceInfo)

    state = new State($x$, $f(x)$, $\nabla f(x)$, $h$,
                      convergenceInfo = newCInfo)
}
}
\end{lstlisting}
Thus all that any concrete implementation of \lstinline{FoMin} needs to do
is implementing a few methods and data types:
\begin{itemize}
    \item \lstinline{History}: representing information from past
    iterations needed to take decisions at the current iteration.
    \item \lstinline{History}'s initialization and management via
    \lstinline{initialHistory} and \lstinline{updateHistory}.
    \item Possibility of modifying the objective, e.g.~by adding regularization
    via \lstinline{adjustFunction}.
    \item Choosing a descent direction and a step size via \lstinline{chooseDescentDirection}
    and \lstinline{determineStepSize}.
    \item Updating the $x$ using \lstinline{takeStep}.
    \item Implementing convergence conditions in \lstinline{convergenceCheck}.
\end{itemize}
\subsection{Implementations of {\normalfont\lstinline{FirstOrderMinimizer}}}\label{subsec:fom-implementation}
With this framework several algorithms can be implemented.
    \begin{enumerate}
        \item Stochastic Gradient Descent which is history-less implementing
        a simple update:
        \begin{equation}
            x_t = x_{t-1} - \eta_t\nabla f(x_{t-1})
        \end{equation}
        where $\eta_t$ is either a constant learning rate or a learning
        rate decaying via a power law (in the number of iterations)
        specified by the user.
        \item (a flavor of) Adagrad~\cite{duchi-adaptative} which adjusts the learning rate
        for each component of $x_t$ dynamically incorporating knowledge
        of the observed data to perform more informative choices at each
        \lstinline{takeStep}.
        Concretely, $\tilde f$ can be adjusted adding $l_1$ and $l_2$
        regularizations:
        \begin{equation}
            f(x) = \tilde f(x) + \alpha_1\|x\|_1 + \frac{\alpha_2}{2}\|x\|_2^2.
        \end{equation}
        History is accumulated in a vector:
        \begin{equation}
            h_{t,i} = \begin{cases}
                          \sum_{s=1}^{t-1}(\nabla_i f(x_s))^2 & \textrm{if $t\le m$}, \\
                          \left(1-\frac{1}{m}\right) h_{t-1, i} + \frac{1}{m}(\nabla_i f(x_t))^2
                          & \textrm{if $t>m$},
            \end{cases}
        \end{equation}
        where $m$ is a parameter controlling the memory length in accumulating
        history.
        The descent direction is just $\nabla f(x_{t-1})$ and the learning
        rate is just a constant $\eta$;
        however steps are taken adaptively.
        First one builds a direction-stretch $\sigma_t$ vector:
        \begin{equation}
            \sigma_{t,i} = \sqrt{h_{t-1,i} + (\nabla_i f(x_{t-1}))^2 + \delta},
        \end{equation}
        where $\delta$ is usually a small parameter supplied by the user which
        avoids division by $0$.
        Then a tentative step without $l_1$-regularization is taken:
        \begin{equation}
            \tilde x_{t,i} = \frac{\sigma_{t,i}x_{t-1} + \eta\nabla_i f(x_{t-1})}{
            \sigma_{t,i} + \eta\alpha_2
            }.
        \end{equation}
        Finally a step with $l_1$-regularization is taken:
        \begin{equation}
            x_{t,i} = \begin{cases}
            0 & \textrm{if $|\tilde x_{t,i}| < \frac{\eta\alpha_1}{\sigma_{t,i}}$}\\
                          \tilde x_{t, i} - \frac{\eta\alpha_1}{\sigma_{t,i}}\sign{\tilde x_{t,i}}
                          &\textrm{otherwise}.
    \end{cases}
        \end{equation}
        \item LBFGS: In this Quasi-Newton method~\cite[Chap.~8]{NoceWrig06} the descent direction at iteration $t$
        is of the form:
        \begin{equation}
          w_t = -H_t\nabla \tilde f(x_{t-1}),
    \end{equation}
where $H_t$ is a matrix approximating the inverse of the Hessian $\nabla^2\tilde f(x_{t-1})$ \emph{along the
        gradient direction} $\nabla \tilde f(x_{t-1})$.
In reality one does not need to store the whole $H_t$ but just $m$, being a user-defined parameter, previous gradients.
The details of computing $H_t$ and freeing computational resources are handled by the implementation of
\lstinline{History}.
To determine the step size one uses any line search algorithm, see~\cite[Chap.~3]{NoceWrig06}.
At the moment of writing implementations for a Strong Wolfe Line Search and a Back Tracking Line Search are provided.
        \item OWLQN: this is an \emph{orthant-wise} Quasi Newton method proposed in~\cite{andrew-scalable}.
This method is built on top of LBFGS for problems which have an additional $l_1$-regularization.
As the $l_1$-regularization is non-smooth with corner points, this method aims at speeding the convergence
properties of LBFGS.
From an implementation point of view one just needs to add some projection-like operations on orthants as explained
in~\cite{andrew-scalable}.
In~\cite{andrew-scalable} it is shown that OWLQN  can work well with large scale log-linear models.
However, as observed by the authors of the breeze package on general problems the algorithm might fail to converge
and we are not aware of theoretical guarantees regarding its convergence.
     \end{enumerate}
    \section{Losses and Models}\label{sec:losses-models}
    \subsection{Regularized Optimal Transport}\label{subsec:reg-opt}
For an introduction and review of Optimal Transport we refer the reader to~\cite{cuturi_book, santambrogio_book};
here we will follow the notations of~\cite[Sec.~2]{schioppa-neural}.
Specifically, we focus on the \emph{discrete} formulation of the \emph{entropy-regularized dual}
problem, with the following loss to \emph{maximize}:
\begin{equation}\label{eq:opt-problem}
    \mathcal{L}=\frac{1}{N_{\Xs}}\sum_{i=1}^{N_\Xs} u_i +
    \frac{1}{N_{\Ys}}\sum_{j=1}^{N_\Ys} v_j -
    \frac{\varepsilon}{N_\Xs N_\Ys}\sum_{i=1}^{N_\Xs}\sum_{j=1}^{N_\Ys} \exp\left(
    \frac{u_i + v_j - c_{i,j}}{\varepsilon}
    \right),
\end{equation}
where $u=\{u_i\}_{i=1}^{N_\Xs}$, $v=\{v_j\}_{j=1}^{N_\Ys}$ are variables to optimize (called the potentials),
$c_{i,j}$ is the (pre-computed) cost matrix and $\varepsilon$ is the regularization strength (which we want small).
In our case both $N_\Xs$ and $N_\Ys$ are large making necessary to distribute the potentials across
node clusters;
the cost matrix can be quite large, in general a dense $N_\Xs \times N_\Ys$-matrix that Spark will partially
cache in memory and partially leave on disk.
\par
\begin{figure}[ht]
        \caption{Our computational approach to distribute the
        optimal transport loss.
        We combine the potentials inside the
        same distributed vector and partition costs compatibly.}
        \label{fig:opt-computation}
    \includegraphics[height=6cm,width=10cm]{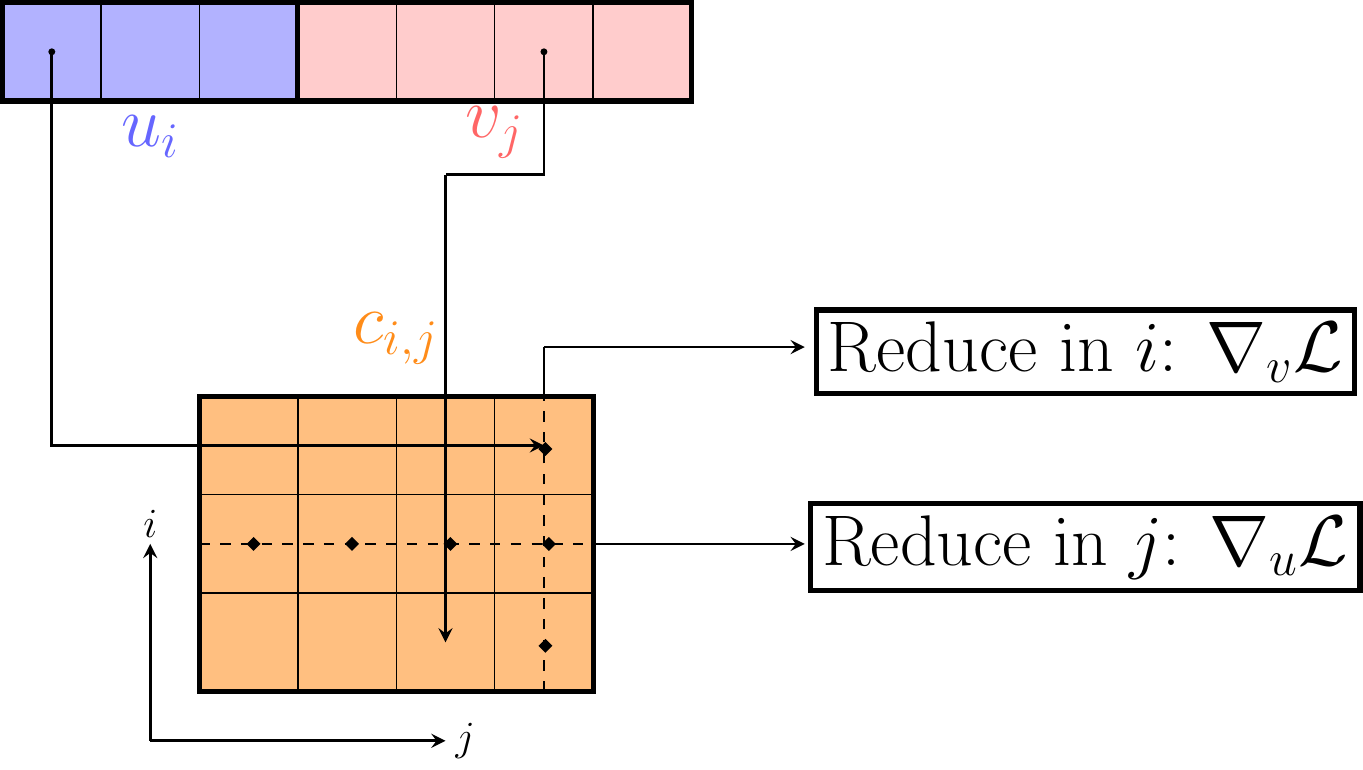}
    \end{figure}

 In Figure~\ref{fig:opt-computation} we describe the computational approach
that we use.
The potentials $u$, $v$ are combined together in the same \lstinline{FomDistVec};
each partition of $u$ (resp.~$v$) is replicated by the number of partitions of $v$
(resp.~$u$);
the cost matrix $c$ is partitioned by a grid allowing to compute all the regularization
terms involving triplets $(u_i, v_j, c_{i,j})$.
The total loss $\mathcal{L}$ is computed using a \lstinline{treeAggregate}
on the grid;
the gradient $\nabla_u\mathcal{L}$ (resp.~$\nabla_v\mathcal{L}$)
is computed doing a reduction (via a sum) in $j$ (resp.~$i$).
\subsection{Linear Models}\label{subsec:lin-models}
\begin{figure}[ht]
        \caption{Example of distributed linear scoring with
        $m_{\mathcal{E}} = 3$ and $m_{\mathcal{F}} = 4$.
        The example $e$ does not have features for the partitions
        $i=1, 3$.}
        \label{fig:linear-scorer}
    \includegraphics[height=6cm,width=10cm]{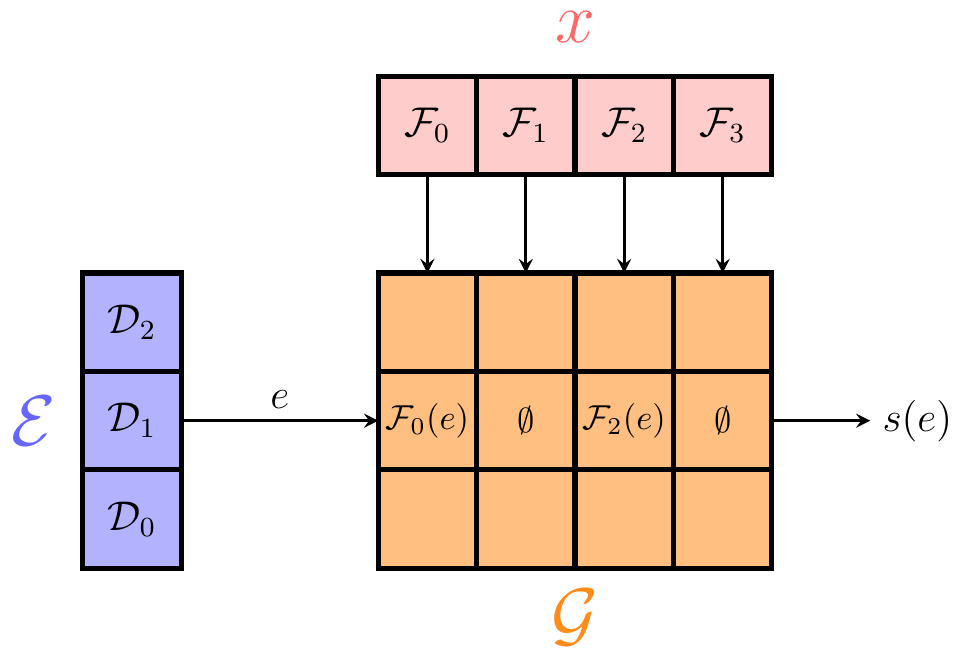}
    \end{figure}

In Figure~\ref{fig:linear-scorer} we describe the approach that we
follow to score linear models.
For simplicity, let us focus on the case that a single score needs
to be produced for each training example.
Our approach mixes \emph{data parallelism} with \emph{model parallelism}.
Let $x$ denote the distributed feature vector where the features,
indexed by $\alpha$, are partitioned into sets $\{\mathcal{F}_i\}_{i=0}^{m_\mathcal{F}-1}$;
let $\mathcal{D}$ denote the data-set of training examples
and partitioned into sets $\{\mathcal{D}\}_{j=0}^{m_\mathcal{E}-1}$.
Let $e\in\mathcal{D}_j$ be an example;
the features of $e$ are stored into a sparse vector
$\{f_\alpha\}_\alpha$ (i.e.~each $f_\alpha\ne0$) and
we denote the indices $\alpha$ such that $\alpha\in\mathcal{F}_i$
by $\mathcal{F}_i(e)$.
While Spark-ML would broadcast $x$ across all the data points, we
cannot do that.
Neither we have at our disposal a centralized \emph{parameter server}
that we can use to fetch weights from for each example.
The approach we take is thus to partition both $x$ and the features of each
example by creating a computational grid $\mathcal{G}$ such that the
cell $(j, i)$ contains a copy of $\{x_\alpha\}_{\alpha\in\mathcal{F}_i}$
and all the features $\{\mathcal{F}_i(e)\}_{e\in\mathcal{D}_j}$.
Thus we can compute the score $s(e)$ in a distributed way via the
formula
\begin{equation}
    s(e) = \sum_{i: \mathcal{F}_i(e)\ne\emptyset}
    \sum_{\alpha\in\mathcal{F}_i(e)}f_\alpha x_\alpha,
\end{equation}
where the outer sum is implemented as a \lstinline{reduceByKey}
in $e$.
\subsection{Factorization Machines}\label{subsec:fact-machines}
Factorization Machines~\cite{rendle-bpr, rendle-fm} produce the scores via pairwise
interactions across all the distinct features.
In this case $x$ if of the form $\{(x_\alpha^k)_{\alpha}\}_{k=0}^{m_L-1}$,
where $m_L$ is the \emph{latent dimension} assumed much smaller than the possible
range of $\alpha$;
in our case we represent $x$ via an \lstinline{FDSDV}.
Now using the notation of Subsection~\ref{subsec:lin-models} we can write
\begin{equation}\label{eq:fm-score}
    s(e) = \sum_{\alpha,\beta \atop \alpha\ne\beta}
    \sum_{k=0}^{m_L-1}f_\alpha f_\beta x_\alpha^k x_\beta^k.
\end{equation}
Distributing this computation can be reduced to the case of linear models
by rewriting $s(e)$ as:
\begin{equation}
    \begin{split}
s(e) &= \frac{1}{2}\sum_{k=0}^{m_L-1}\Biggl(\mskip 10mu
\underbrace{\sum_\alpha f_\alpha x_\alpha^k}_{t_{\mathrm{l}}^k}
\mskip 10mu
\Biggr)^2 - \underbrace{\sum_\alpha (x_\alpha^k)^2 f_\alpha^2}_{t_{\mathrm{q}}^k}\\
        &=\frac{1}{2}\sum_{k=0}^{m_L-1}(t_{\mathrm{l}}^k)^2 - t_{\mathrm{q}}^k.
\end{split}
\end{equation}
Now the vectors $(t_{\mathrm{l}}^k)_k$ and $(t_{\mathrm{q}}^k)_k$
can be computed using the distributed approach presented in the previous section.
\subsection{Regression and Back-propagation}\label{subsec:reg-back-prop}
\begin{figure}[ht]
        \caption{Representation of a gradient computation when $s(e)$ is
        produced by a linear model.
        The dashed lines refer to the ``forward'' pass while the
        solid lines to the ``backward'' pass, i.e.~the gradient computation.}
        \label{fig:regression-backprop}
    \includegraphics[height=6cm,width=12cm]{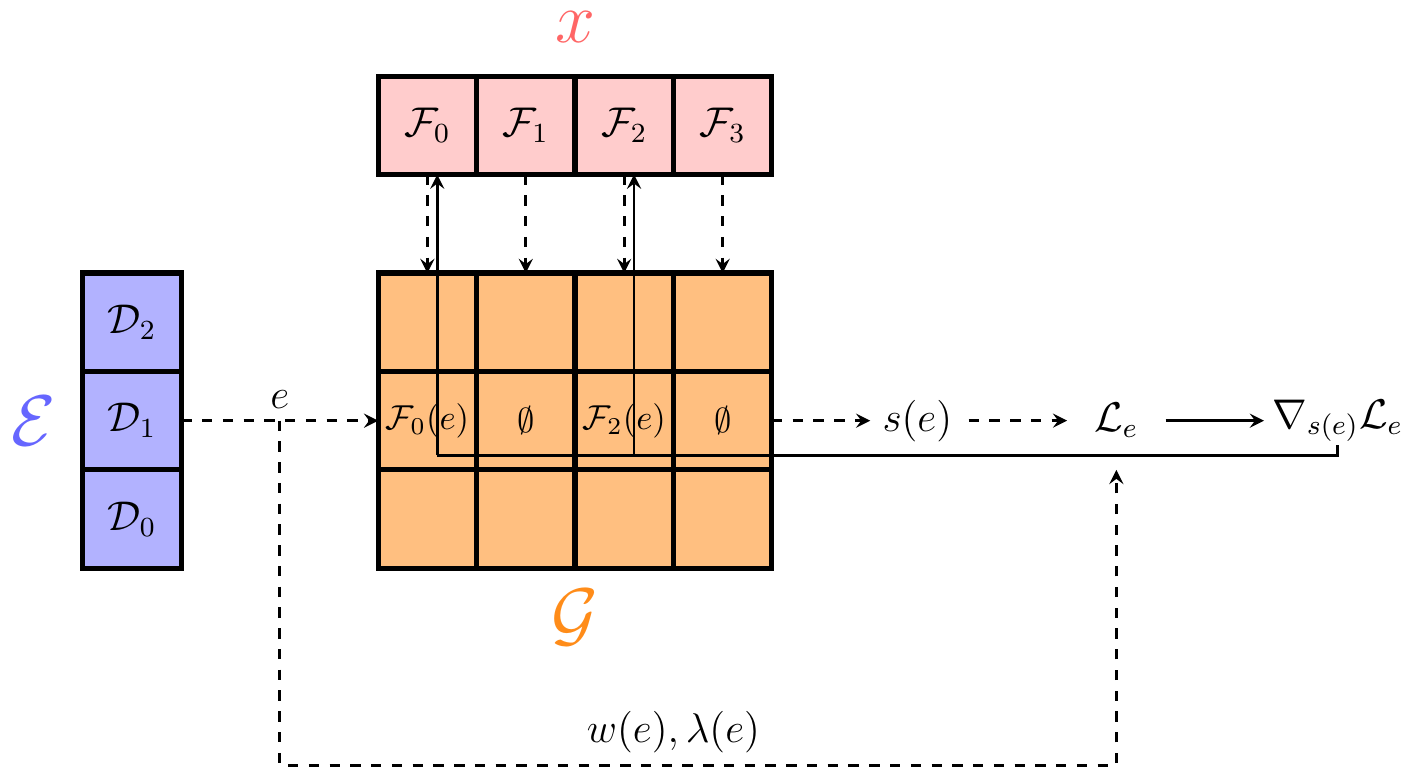}
    \end{figure}

Up to now we have just discussed the construction of model scores $s(e)$;
such a score will enter, together with a label $\lambda(e)$ and a weight
$w(e)$ the computation of the loss
$\mathcal{L}_e=w(e)\mathcal{L}(s(e),\lambda(e))$.
Computing the scores and the loss is what in the Deep Learning Literature
terminology is usually called the ``forward'' pass;
we now describe the so-called ``backward'' pass, i.e.~the computation
of the gradient:
\begin{equation}
    \nabla_x\mathcal{L} = \frac{1}{\sum_{e\in\mathcal{E}}w(e)}
    \sum_{e\in\mathcal{E}}\nabla_x\mathcal{L}_e.
\end{equation}
For the following discussion please refer to Figure~\ref{fig:regression-backprop};
for the moment we just focus on the case in which $s(e)$ is produced by a linear
model.
While the computational approach is generic in $\mathcal{L}_e$, in the library we have just
implemented the $l_2$-loss, the quantile loss and the logistic loss for
binary classification.
\par Coming back to the computation of $\nabla_x\mathcal{L}$, the chain rule
gives:
\begin{equation}
    \nabla_x\mathcal{L}_e = \nabla_{x}s(e)\mskip 3mu\cdot\mskip 3mu
    \nabla_{s(e)}\mathcal{L}_e,
\end{equation}
$\nabla_{s(e)}\mathcal{L}_e$ being just a scalar;
specializing to an index $\alpha$ such that $f_\alpha\ne0$ we get:
\begin{equation}
    \nabla_{x_\alpha}\mathcal{L}_e = f_\alpha
    \nabla_{s(e)}\mathcal{L}_e.
\end{equation}
Thus for each $e$ we just need to keep track those $i$'s for which
$\mathcal{F}_i(e)\ne\emptyset$ and pull back $\nabla_{s(e)}\mathcal{L}_e$
to those cells $(j(e), i)$ on $\mathcal{G}$ where $\mathcal{D}_{j(e)}$ is the data-partition
containing $e$;
finally we aggregate across the examples the contributions to each
gradient component obtaining $\nabla_x\mathcal{L}$ as in Figure~\ref{fig:regression-backprop}.
\subsection{Multi-classification and Back-propagation}\label{subsec:multi-class-back-prop}
We have also implemented multi-classification models where $\mathcal{L}$ is a
softmax followed by a logloss.
We allow multi-label targets with different weights for each target
but we will not spell it out in the computational details for the sake
of exposition.
Now $x$ is of the form $\{(x_\alpha^k)_{\alpha}\}_{k=0}^{m_C-1}$,
$m_C$ being the number of possible classes;
in our case we represent $x$ via an \lstinline{FDSDV}.
Now for each $k$ we obtain a score $s^k(e)$ and the gradient of the
loss takes the form:
\begin{equation}
\nabla_{x_\alpha^k}\mathcal{L}_e =
    \frac{\partial s^{k}(e)}{\partial x_\alpha^k}
    \nabla_{s^{k}(e)}\mathcal{L}_e = f_\alpha\nabla_{s^{k}(e)}\mathcal{L}_e;
\end{equation}
this computation can be handled similarly as in Subsection~\ref{subsec:reg-back-prop}
as the added complexity just amounts to handling the index $k$, e.g.~in computing
the gradient of the loss with respect to all the possible $k$'s.
\subsection{Ranking with Negative Sampling and Back-propagation}\label{subsec:rank-neg-sample}
As an application of Factorization Machines we consider the ranking problem
with \emph{implicit feedback} following~\cite{rendle-bpr, rendle-fm}.
Whenever a user $u$ makes a query / search $q$ it receives a list of
recommended items;
each time an item is selected we form a positive example as a triplet
$(u, q, i)$.
We will denote the set of positive examples by $\mathcal{E^+}$.
\par For the moment we set modeling aside and assume that for each user / search pair
$(u, q)$ and each item $k$ we can associate a score $s(u, q; k)\in\real$;
the higher the score the more likely is $k$ to be selected by the user.
In the implicit feedback setting we want to optimize $s$ simply to give
higher score to selected items compared to non-selected ones.
Mathematically, to each $(u, q; i)$ we associate a probability distribution
$\mathcal{N}(u, q; i)$ from which we sample ``negative'' items
$j\sim\mathcal{N}(u, q; i)$;
this framework is quite general, as for example the query $q$ and $i$
can narrow down the negative items eligible for sampling.
\par Let $\sig(x) = 1/(1+\exp(-x))$ be the sigmoid function;
then the loss function for the example $(u, q; i)$ becomes:
\begin{equation}
    \mathcal{L}_{(u, q; i)} = -\expectation\log\sig(s(u, q; i)
    - s(u, q; j)): j\sim\mathcal{N}(u, q; i).
\end{equation}
In practice we fix a parameter $n_{\mathcal{N}}$ and we draw
a sample $\mathcal{N_S}(u, q; i)\sim\mathcal{N}(u, q; i)$ of size
$n_{\mathcal{N}}$ and compute:
\begin{equation}
    \mathcal{L}_{(u, q; i)} = -\frac{1}{n_{\mathcal{N}}}
\sum_{j\in \mathcal{N_S}(u, q; i)}\log\sig(s(u, q; i)
    - s(u, q; j)).
\end{equation}
Our implementation is a bit more general;
we represent $(u, q; i)$ just as an example $e$ (allowing more generality
in the way features are encoded) and we let the user be able to specify
a \lstinline{NegativeSampler} class to generate the negative samples
$\mathcal{N_S}(u, q; i)$.
\par
\begin{figure}[ht]
        \caption{Representation of ranking computations using Factorization
        Machines and implicit feedback.
        While only example features are need to produce the scores
        $\mathcal{S}^+$, $\mathcal{S}^-_{0/1}$, the losses use joins
        $\mathcal{J}^+$ (resp.~$\mathcal{J}^-_{0/1}$) of
        $\mathcal{S}^+$ (resp.~$\mathcal{S}^-_{0/1}$) with
        $\mathcal{T}^+$ (resp.~$\mathcal{T}^-_{0/1}$).}
        \label{fig:ranking-loss}
    \includegraphics[height=12cm,width=9cm]{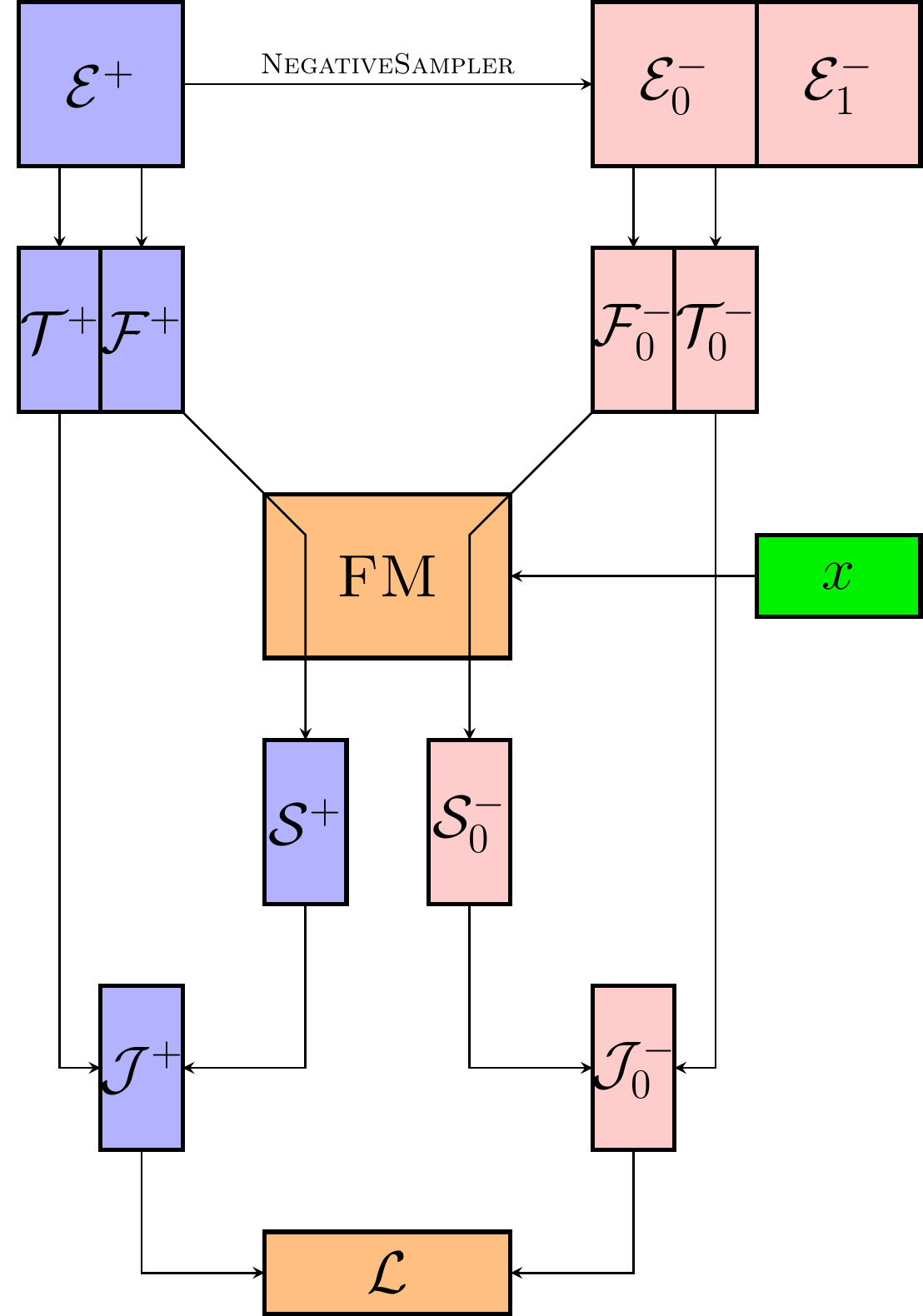}
    \end{figure}

 In Figure~\ref{fig:ranking-loss} we describe the computational
model for the loss.
The data consists of positive examples $\mathcal{E}^+$ and the
\lstinline{NegativeSampler} produces batches of negative examples
$\{\mathcal{E}^-_b\}_b$;
each computation is looped on the batch index $b$ (and gradients /
losses are aggregated in $b$).
From the positive examples $\mathcal{E}^+$ we extract the labels
$\mathcal{T}^+$ and the features $\mathcal{F}^+$; similarly from
each $\mathcal{E}^-_b$ we extract $\mathcal{T}^-_b$ and
$\mathcal{E}^-_b$.
The data structure for the labels contains a ``link'' field to
be able to identify which negative examples have been sampled for
a specific positive example.
The Factorization Machine model is then used to score positive
and negative examples producing $\mathcal{S}^+$, $\mathcal{S}^-_b$;
finally combining the labels and using the ``link'' field one gets the
loss $\mathcal{L}$.
\par Let us spend a few words on the gradient computation.
We can compute the gradient looking at the individual losses
$\mathcal{L}(e^+, e^-)$ for each pair of positive / negative items.
To get $\nabla_x\mathcal{L}(e^+, e^-)$ we first need two
derivatives $\nabla_{s(e^+)}\mathcal{L}(e^+, e^-)$,
$\nabla_{s(e^-)}\mathcal{L}(e^+, e^-)$;
in the backward step, one derivative will flow back to the
positive features, the other one to the negative features.
Let us have a look at the derivative of the scores produced by
the Factorization Machine: from \eqref{eq:fm-score} we get:
\begin{equation}
    \nabla_{x_\alpha^k}s(e) = \sum_{\beta\ne\alpha}f_\alpha f_\beta x_\beta^k;
\end{equation}
so during the forward step we just need compute the vector $v(e)$ where
\begin{equation}
    v(e)^k = \sum_\beta f_\beta x_\beta^k
\end{equation}
that is used during the backward step to compute the gradient $\nabla_{x_\alpha^k}s(e)$.
\subsection{A few comments about batching}\label{subsec:comm-batching}
For large data-sets we allow the user to speed-up computations by
splitting the data into batches.
    Here we think of a few number of batches, i.e.~of batches
    with many examples, or ``macro''-batches.
\par For models with many parameters too-small macro-batches can lead
    to over-fitting.
    During line searches this can manifest itself into taking too large steps.
    This is why we allow, optionally, to use a different batch for
    choosing the step-size than the batch used for choosing the
    descent direction.
    Together with implementation of regularization losses this provides a
    simple tool to prevent over-fitting.
\par We think that a better theoretical understanding of the behavior
    of Quasi-Newton methods that use ``macro''-batches is an interesting
    topic for further research.
    As well we leave for further work to add more tools to the library to
    prevent over-fitting.
\section{Scalability Experiments}\label{sec:scalability-exp}
    We ran 3 ``scalability'' experiments to gain insight into the
library performance.
    These experiments \emph{were not setup as benchmarks};
    for example the configuration of the Spark cluster was
    standardized and not much tuned, loosely following this
\href{https://blog.cloudera.com/how-to-tune-your-apache-spark-jobs-part-2/}{blog}
\footnote{https://blog.cloudera.com/how-to-tune-your-apache-spark-jobs-part-2/}.
    Similarly, timings were taken mainly to understand the cost of loading data
    and evaluating gradients across iterations.
    Finally, ML problems were chosen on \emph{synthetic data} to
create an unfavorable situation where features are sparse but yet one would
need to learn many parameters to fit a model.
Keys for the sparse features were selected using uniform distributions to
make the problem kind of unfavorable from a scalability viewpoint.
This is in contrast with the ``manifold hypothesis''~\cite{arjovsky_principled} that assumes
data living on a low-dimensional sub-manifold of the highly dimensional
feature space.
For this reason we avoid handling high cardinality features via the hashing trick
(like in Vowpal Wabbit) or by deleting infrequent keys (like in some Parameter
Server implementations on GitHub, e.g. \href{https://github.com/dmlc/ps-lite}{this}
\footnote{https://github.com/dmlc/ps-lite}).
We note however that in the current library dimensionality reduction can be achieved
either by adding $l^1$-regularization or by using OWLQN as an optimizer.
    \par We set up the Spark Cluster to run with dynamic allocation
with preemptible workers and speculation enabled.
This implies that the number of active cores and the memory claimed from
    the cluster can change dynamically during execution.
We did however cap the maximal number of executors to 400 with 2 cores
    per executor and 8GB of memory per executor.
Note that executors are \textbf{not} nodes:
depending on the number of cores per node one might need a cluster of
variable size, e.g.~200 nodes or 50 nodes.
This is a rather modest amount of resources compared to~\cite{langford-all-reduce} (which
does not specify nodes setting) or~\cite{li-param-server}.
    Finally note that these settings
imply that memory usage for our jobs cannot exceed 3.2TB.
    For the optimal transport problem we tightened these constraints
    to 100 executors so at most 800GB of memory usage.
\subsection{Modeling experiments}\label{subsec:model-exp}
\par The first two experiments concern modeling tasks on synthetic
    data.
We do so in order to focus on scalability, as we make the data
generating process \emph{known}, or in other words ``a best''
model is known.
\par The first experiment is an $l_2$-regression;
    the cardinality of the model vector $w=\{w_i\}_i$ is $10^9$
    with each $w_i$ sampled uniformly from $[0, 1]$ independently
    of the others and $w$ divided into 100 partitions.
    For each example $e$ we choose a set $\mathcal{I}(e)$ of 30
    indices and for each $i\in\mathcal{I}(e)$ we sample the
    feature $f_i(e)$ uniformly from $[-1.0, 1.0]$ and independently
    from the other features.
    The response / label for $e$ is \emph{deterministic}:
    \begin{equation}
        y(e) = \sum_{i\in\mathcal{I}(e)}f_i(e) w_i.
    \end{equation}
    We construct 5 batches each one consisting of $10^8$ examples
    partitioned into 100 partitions.
    \par For the second experiment we consider a multi-classification
    task with 10 categories.
    The model consists of parameters $w=\{w^k_i\}_{k, i}$ with
    $k\in\{0,\ldots, 9\}$ and $i\in\{0,\ldots, 5\cdot 10^7-1\}$ where each
    $w^k_i$ is sampled uniformly in $[0, 1]$ and independently from
    the others;
    we still split $w$ into 100 partitions.
    For each example $e$ we choose a set $\mathcal{I}(e)$ of 100
    indices and for each $i\in\mathcal{I}(e)$ we sample the
    feature $f_i(e)$ uniformly from $[-1.0, 1.0]$ and independently
    from the other features.
    We then generate the label for $e$ extracting $k$ with probability
    \begin{equation}
        p_k(e) \propto \exp\left(
            \sum_{i\in\mathcal{I}(e)}w^k_i f_i(e)
        \right).
    \end{equation}
    We construct 5 batches each consisting of $2.5\cdot 10^8$ examples
    divided into 250 partitions.
    Note that even if the data generating process is known, in this
    case it is not \emph{deterministic} so one cannot expect to
    obtain a logloss of $0$.
In both cases we use LBFGS for loss minimization.
    \par
\begin{figure}[ht]
        \caption{Experiment 1: Over-fitting on just one batch
        and regular batching;
        data reported up to the stopping iterations or a maximum of 10}
        \label{fig:experiment1}
    \includegraphics[height=8cm,width=12cm]{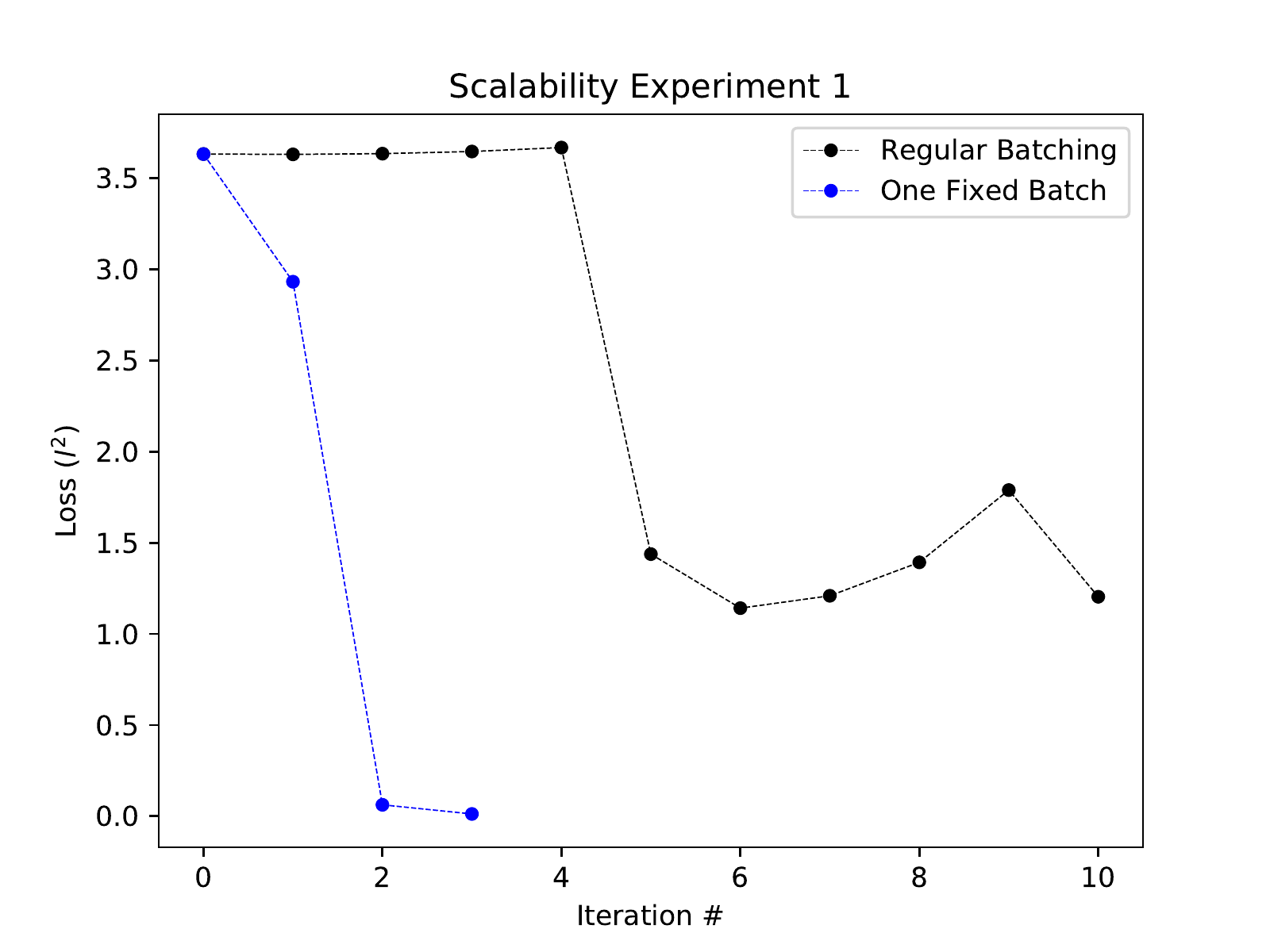}
    \end{figure}

 In the first experiment we can perfectly over-fit
    using a single batch in $4$ iterations.
    On the other hand if we load a new batch after each iteration
    it takes 5 iterations (i.e.~a full pass over the data) to see
    a significant decrease in the loss, see Figure~\ref{fig:experiment1}.
    \par
\begin{figure}[ht]
        \caption{Experiment 2: Different approaches to batching
        in a multi-classification problem;
        data reported up to the stopping iterations or a maximum of 10}
        \label{fig:experiment2}
    \includegraphics[height=8cm,width=12cm]{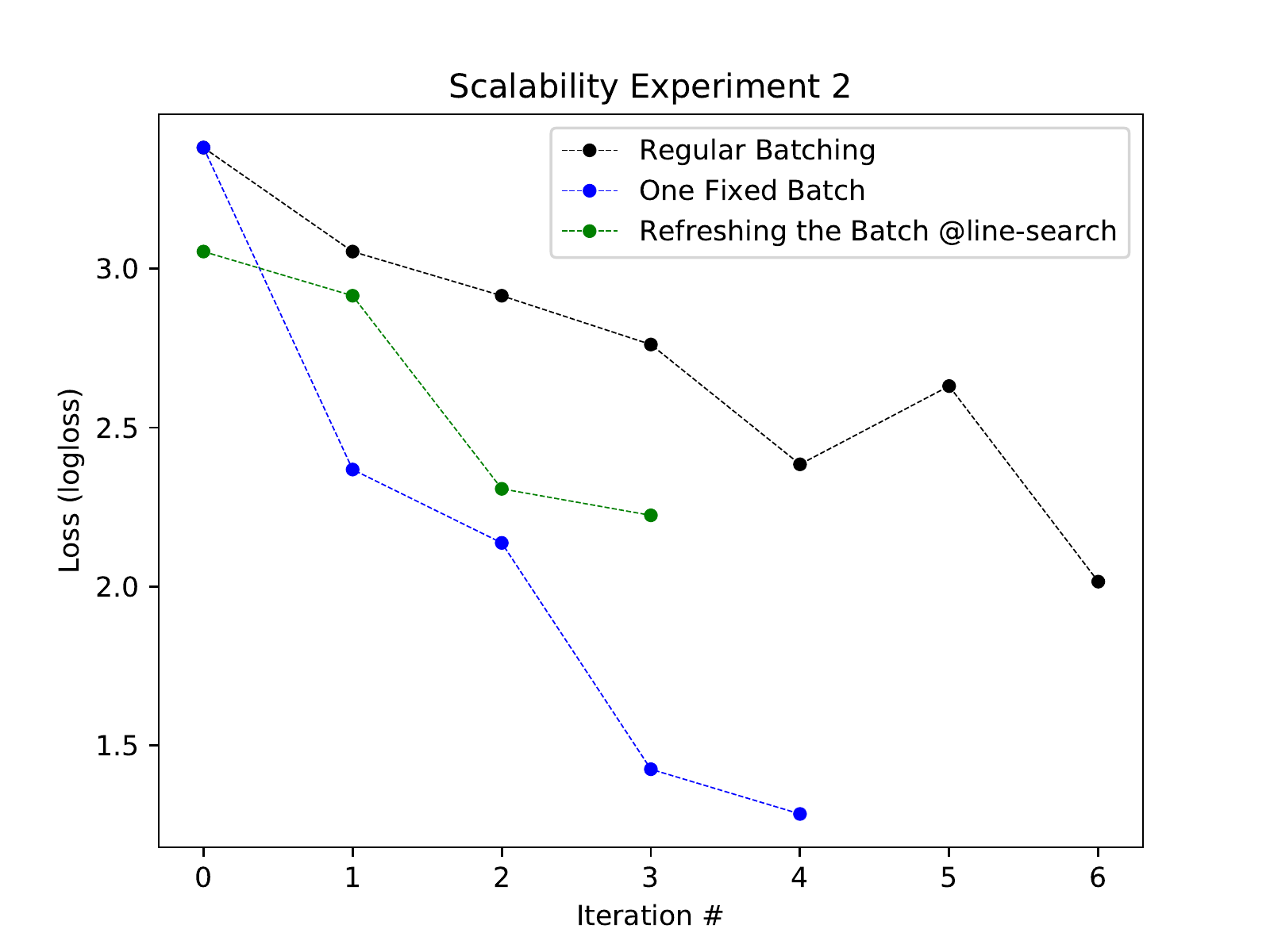}
    \end{figure}

 In the second experiment, see Figure~\ref{fig:experiment2},
    one benefits from using a
    different batch during the line search than the one used to
    compute the gradient.
    For example, after 4 iterations the first approach yields a loss
    20\% lower than the one which keeps the same batch (and loads a new
    batch just for the next iteration).
\par For the first experiment the time to load and cache a batch
of features is $8.17 \pm 1.86$ minutes and the time of each function
value / gradient computation is $5.54 \pm 1.55$ minutes.
For the second experiment the time to load and cache a batch
of features is $30.19 \pm 5.82$ minutes and the
the time of each function value / gradient computation is
$1.77 \pm 0.20$ hours.
In the second experiment we suspect the slow down is driven
by network communication and, with the limited amount of resources
we use,
might be reduced by adding to the library a function to aggregate
gradients across ``macro batches''.
  \subsection{Optimal Transport}\label{subsec:opt-exp}
\begin{figure}[ht]
        \caption{Experiment 3: Convergence of the Optimal Transport
        Loss with Entropic Regularization.
        Note the negative sign as we minimize $-\mathcal{L}$ in
        \eqref{eq:opt-problem}}
        \label{fig:experiment3_funvalue}
    \includegraphics[height=8cm,width=12cm]{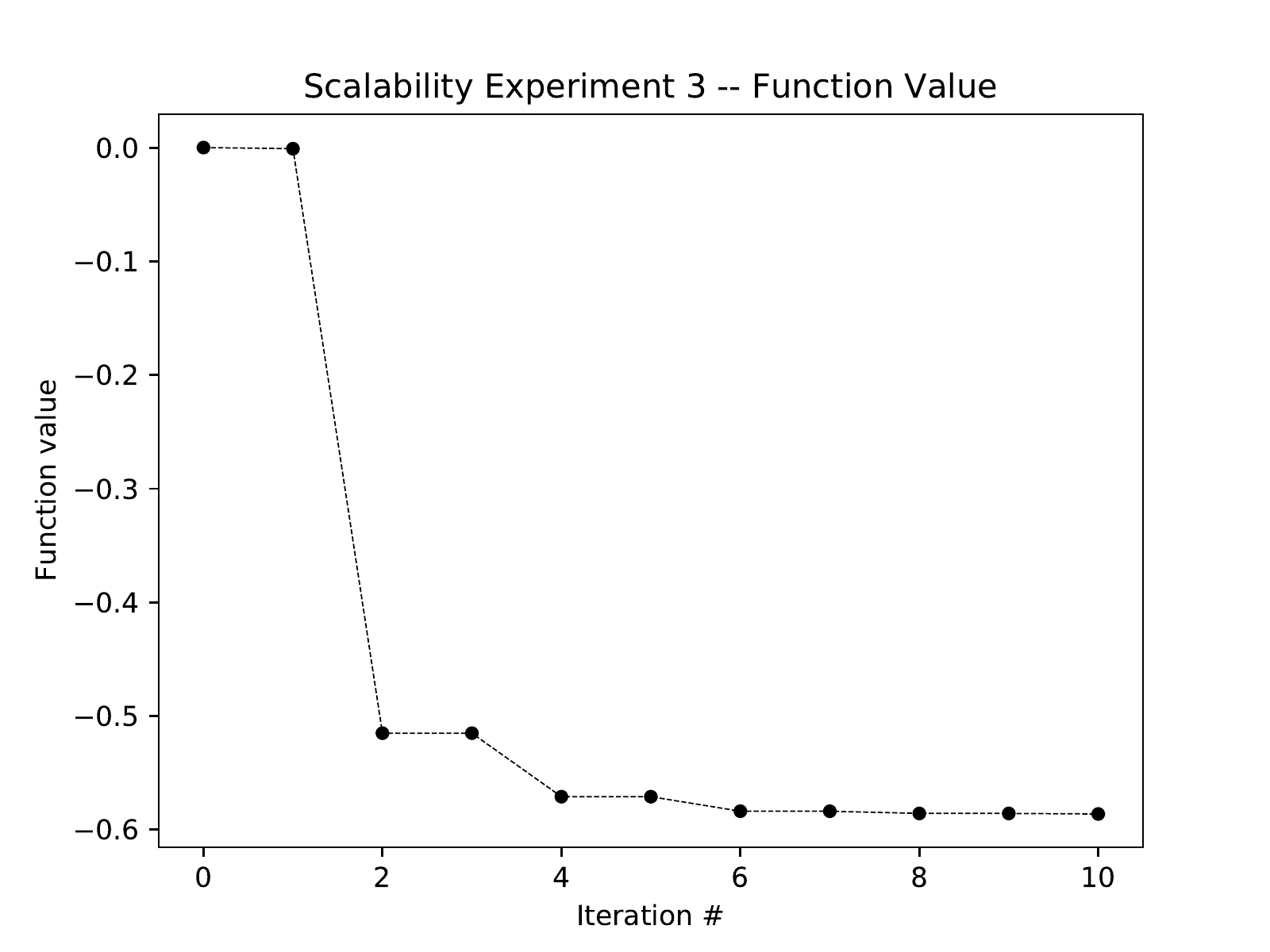}
    \end{figure}

\begin{figure}[ht]
        \caption{Experiment 3: Convergence of the
        gradient norm in the Optimal Transport
        Problem with Entropic Regularization}
        \label{fig:experiment3_gradnorm}
    \includegraphics[height=8cm,width=12cm]{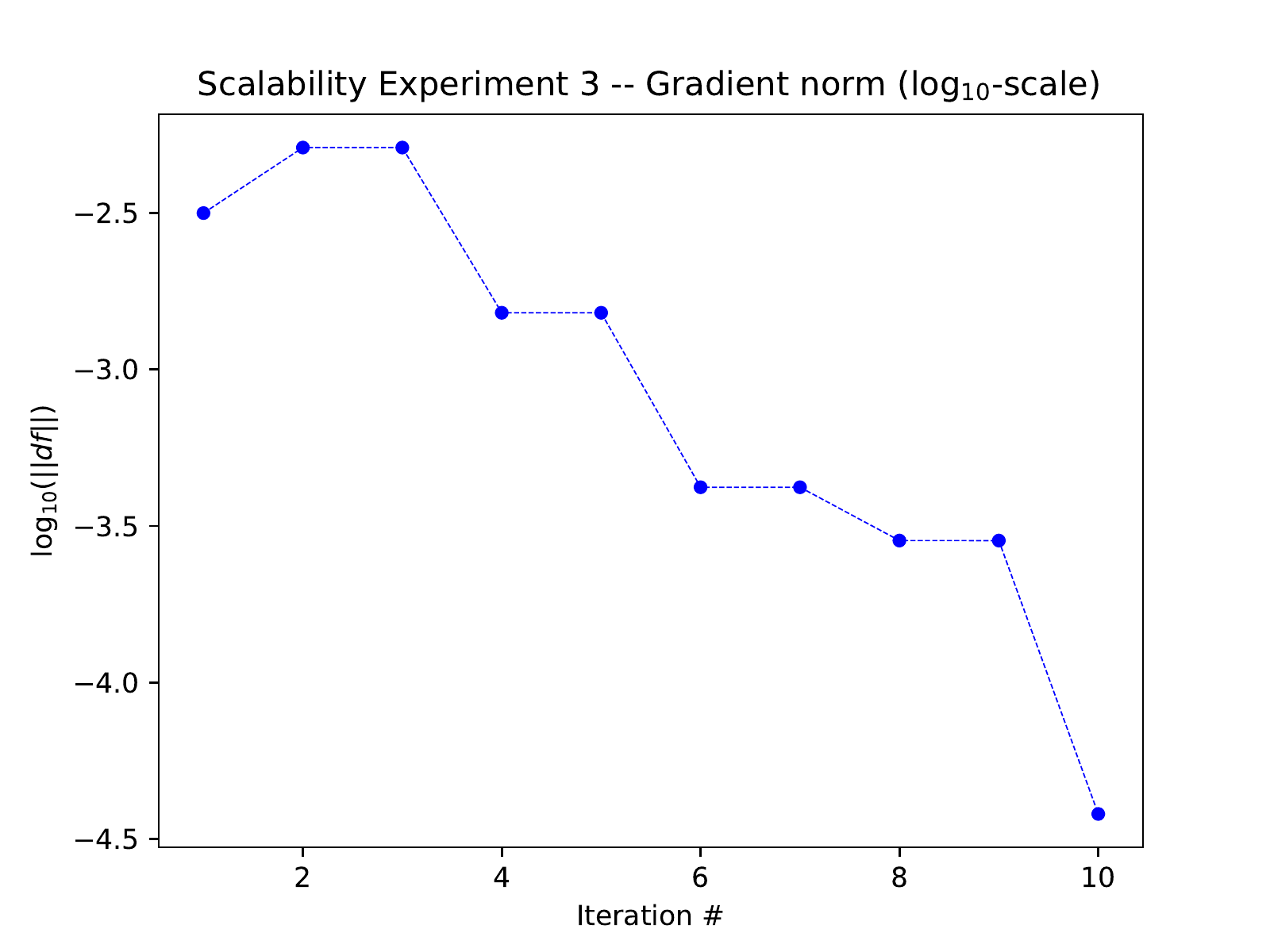}
    \end{figure}

    For the optimal transport problem~\eqref{eq:opt-problem}
    we choose $\Xs=\Ys=\real^{55}$ with $N_{\Xs}=N_{\Ys}=2.5\cdot 10^5$
    and $\varepsilon=0.1$.
    The source distribution is the uniform one on the unit ball of
    $\real^{55}$.
    For the destination distribution we randomly sample $20$
    dimensions $i$ and sample uniformly from the 40 balls
    of radius $1/2$ obtained
    by shifting the origin by $1/2$, $-1/2$ across the direction $i$.
    The vectors $u$ and $v$ are split into $50$ partitions yielding
    $25\cdot 10^3$ partitions for the cost.
    We set the convergence criterion to having the gradient norm
    less than $0.5\cdot 10^{-4}$ which we achieve in $10$ iterations
    using LBFGS~see Figures~\ref{fig:experiment3_funvalue},
    \ref{fig:experiment3_gradnorm}.
    The time of each function value / gradient computation is
    $12.25 \pm 4.24$ minutes.
    Note that the cost matrix considered in our problem is about 156 times
    the one considered by~\cite{genevay_large_opt} as we sample $2.5\cdot 10^5$ points
versus the $20000$ considered in~\cite{genevay_large_opt}.
    \bibliographystyle{alpha}
    \bibliography{the-biblio}
\end{document}